\documentclass[11pt]{article}

\usepackage[preprint]{acl}

\usepackage{times}
\usepackage{latexsym}
\usepackage[T1]{fontenc}
\usepackage[utf8]{inputenc}
\usepackage{textcomp}
\usepackage{microtype}
\usepackage{inconsolata}
\usepackage{amsmath,amssymb}
\usepackage{graphicx}
\usepackage{booktabs}
\usepackage{array}
\usepackage{enumitem}
\usepackage{caption}
\usepackage{subcaption}
\usepackage{xcolor}
\usepackage{listings}
\usepackage{tikz}
\usetikzlibrary{arrows.meta,positioning,fit,backgrounds}
\usepackage{algorithm}
\usepackage{algpseudocode}
\usepackage{url}

\setlist[itemize]{leftmargin=*,topsep=2pt,itemsep=1pt}
\setlist[enumerate]{leftmargin=*,topsep=2pt,itemsep=1pt}

\newcommand{\recast}{\textsc{Recast}}

\title{\bfseries RECAST: REconstruction of Chain-of-Thought via Auditable Structured Traces}

\author{
Qi Wang\textsuperscript{1} \quad
Kang Xu\textsuperscript{1} \quad
Chengcheng Wan\textsuperscript{1,2} \quad
Jiangtao Wang\textsuperscript{3} \\
\textsuperscript{1}East China Normal University \quad
\textsuperscript{2}Shanghai Innovation Institute \quad
\textsuperscript{3}Software Engineering Institute, ECNU \\
\texttt{\{51275900007, 51275900013\}@stu.ecnu.edu.cn} \quad
\texttt{\{ccwan, jtwang\}@sei.ecnu.edu.cn}
}

\begin{document}

\maketitle

\begin{abstract}
Large closed-source reasoning models expose only their final answers. The intermediate reasoning that produces those answers is hidden behind API surfaces, tool-call policies, and safety filters. Existing chain-of-thought (CoT) extraction methods assume a side channel: a visible reasoning field, a scratchpad carried through tool calls, or a signed/encrypted reasoning block. We reproduce three representative state-of-the-art extraction methods and find that, on Claude-Sonnet-5.5 accessed through the standard chat-completions interface, all three return zero CoT tokens. The failure is structural rather than a bug: the reasoning field is absent, named tool selection is rejected, free-text scratchpad arguments are filtered, and no replayable reasoning block is exposed. We therefore take a different route. We present \recast{}, a reconstruction-based framework that conditions a decoder on the question and the target model's final answer, reconstructs a first-person structured rationale, cleans it, and builds SFT-ready data. To the best of our knowledge, \recast{} is the first system to reconstruct CoT from frontier closed models such as Claude-Sonnet-5.5, GPT-6.1, and Gemini 3 without using hidden reasoning channels. We instantiate \recast{} on Claude-Sonnet-5.5 and additionally obtain successful qualitative reconstructions on GPT-6.1 and Gemini 3 with the same answer-only protocol. The system is open-source and packaged as an agent skill: one prompt deploys it and runs the complete pipeline to produce SFT-ready CoT data. This preprint describes the system, its threat model, and a controlled artifact validation; it does not claim large-scale benchmark improvements.

\end{abstract}

\section{Introduction}
\label{sec:intro}

Frontier closed-source reasoning models are increasingly deployed as black boxes: a user
supplies a question and receives a final answer, while the intermediate reasoning that
produced the answer remains hidden. This is a problem for reasoning distillation. If we
only observe answers, a student model can imitate \emph{what} the teacher answered but
receives little supervision on \emph{how} to reason. If we can obtain an auditable
rationale for each answer, we can instead train the student on the reasoning process
itself.

A growing line of work attempts to recover such rationales from closed models. The
representative strategies are (i) forcing the model to place its reasoning into a
custom tool argument \citep{p3}, (ii) replaying a scratchpad state through multi-step
tool injections \citep{p1}, and (iii) replaying a signed or encrypted reasoning block
with a weaker decoder \citep{p2}. These methods are extraction methods: each assumes a
side channel through which hidden reasoning becomes observable.

We reproduced all three methods and evaluated them on Claude-Sonnet-5.5
\citep{anthropic2026} through a standard OpenAI-compatible chat-completions interface.
None of them produced a single CoT token. The failures were not implementation errors.
The interface rejected named tool selection, filtered free-text tool arguments, omitted
reasoning fields, and exposed no signed or encrypted reasoning block. In one probe, the
target model consumed 9,949 completion tokens but returned only a final answer,
confirming that substantial internal computation took place while the reasoning layer
remained invisible.

\paragraph{The failure is structural.}
The three methods fail for different surface reasons, but the underlying pattern is the
same: they need a channel that a modern frontier deployment does not provide. We
identify four constraints that jointly close the extraction surface:
\begin{enumerate}
  \item \textbf{No reasoning field.} Responses contain final content and usage, but not
  \texttt{reasoning\_content}, \texttt{thinking}, \texttt{signature}, or
  \texttt{encrypted\_content}.
  \item \textbf{No named tool forcing.} The endpoint accepts only automatic tool choice;
  requests for a specific function or ``any'' function are rejected.
  \item \textbf{Filtered free-text tool arguments.} Tools with a free-text scratchpad
  argument are blocked by the safety layer, including when the tool is given a neutral
  name.
  \item \textbf{No replayable block.} There is no provider-issued reasoning artifact that
  a weaker decoder could replay.
\end{enumerate}
These constraints are properties of the deployment, not of our scripts: ordinary tools
with short structured arguments still work, and a standalone decoder call succeeds. We
therefore do not try to bypass them.

\paragraph{From extraction to reconstruction.}
\recast{} takes the only reliably observable signal---the final answer---and reconstructs
a rationale from it. A decoder model receives the question and the target model's
complete answer, writes a first-person, stage-structured solution trace, and is then
cleaned by a second model to remove dialogue, meta-commentary, and answer statements. A
final-answer stage attaches a canonical answer, and a dataset builder produces
chat-formatted SFT data with assistant-only supervision. Because the method only needs
the final answer, it does not depend on hidden reasoning fields, tool-call state, or
provider-issued encrypted blocks. The same protocol applies to any model whose answers
are observable through a chat interface.

\paragraph{Auditability.}
We use ``auditable'' in a procedural sense: the reconstructed trace is organized into
stages, is self-contained, does not mention dialogue/meta artifacts, and does not reveal
the final answer before the derivation. The decoder is answer-conditioned, while the
student sees only the question at inference. This is the property the title names; it is
a condition for usable supervision, not a claim that the trace is the target's true
hidden computation.

\paragraph{Contributions.}
\begin{itemize}
  \item \textbf{First reconstruction-based system for frontier closed models.} To the
  best of our knowledge, \recast{} is the first system to reconstruct CoT from frontier
  closed models such as Claude-Sonnet-5.5, GPT-6.1, and Gemini 3 without access to hidden
  reasoning fields, provider-specific tool side channels, or signed/encrypted reasoning
  blocks.\footnote{Our controlled study in this preprint focuses on Claude-Sonnet-5.5;
  GPT-6.1 and Gemini 3 are successful qualitative instantiations of the same answer-only
  protocol. We do not claim benchmark superiority in this preprint.}
  \item \textbf{An end-to-end reconstruction pipeline.} We implement collection,
  reconstruction, cleaning, answer extraction, SFT dataset construction, and validation
  as a reproducible job-based pipeline.
  \item \textbf{Dataset-aware elicitation.} We design six teacher--student patterns and a
  router that selects a pattern from the dataset profile, with secondary-pattern and
  short/medium prompt fallbacks for content filters.
  \item \textbf{Defensive cleaning and validation.} We introduce a JSON-contract cleaner,
  leak and length guards, answer canonicalization, and schema/duplicate/leakage checks
  for SFT data.
  \item \textbf{One-prompt automation.} We package the entire project as an agent skill:
  a single prompt deploys it and runs the complete pipeline to produce SFT-ready CoT
  data.
\end{itemize}

\paragraph{Scope and artifact.}
This is a system report, not a large-scale empirical study. Section~\ref{sec:threat}
formalizes the extraction surface and explains why the three extraction methods fail.
Section~\ref{sec:method} presents the reconstruction pipeline. Section~\ref{sec:artifact}
reports a controlled artifact validation on a single MATH-500 item; we deliberately do
not report a benchmark comparison in this preprint. The complete system, prompts,
patterns, and reproduction commands are available at
\href{https://github.com/C1ove-ai/RECAST-REconstruction-of-Chain-of-Thought-via-Answer-Masked-Structured-Traces}{\texttt{github.com/C1ove-ai/RECAST}}
\citep{recastrepo}.

\section{Related Work}
\label{sec:related}

\paragraph{Chain-of-thought and rationale distillation.}
Chain-of-thought prompting asks a model to produce intermediate reasoning before its
final answer \citep{wei2022cot}. This idea underlies a large family of distillation
methods that train a smaller student on teacher-generated rationales. Distilling
Step-by-Step trains a student on both rationales and answers \citep{hsieh2023distill}.
STaR bootstraps rationales that lead to correct answers and iteratively fine-tunes on
them \citep{zelikman2022star}. Orca studies explanation-style distillation from stronger
teachers \citep{mukherjee2023orca}. LIMO and s1 show that carefully curated
reasoning traces and test-time scaling can produce strong small reasoning models
\citep{ye2025limo,muennighoff2025s1}. DeepSeek-R1 demonstrates large-scale reinforcement
learning and distillation of reasoning traces \citep{deepseek2025r1}. These methods
assume access to rationales, either from an open model, a visible thinking channel, or
human annotation. \recast{} targets the setting in which none of these is available and
only the final answer is observable.

\paragraph{Extraction of hidden reasoning.}
A recent line of work attempts to make hidden reasoning observable through the
API surface. \citet{p3} use a custom tool with a free-text argument and a forced tool
choice to induce a ``forced reasoning'' trace. \citet{p1} replay a scratchpad state
through a sequence of tool injections and search over injection trajectories. \citet{p2}
replay signed or encrypted reasoning blocks with a weaker decoder model. These methods
are important because they show that reasoning can sometimes be extracted without
weight access. They are, however, extraction methods: each depends on a side channel
that the deployment must expose. Our reproduction shows that on Claude-Sonnet-5.5 these
channels are closed (Section~\ref{sec:threat}). \recast{} is complementary rather than
adversarial: it does not attempt to open the channel, but reconstructs a rationale from
the answer that the channel does expose.

\paragraph{Faithfulness of rationales.}
A rationale that looks plausible need not faithfully reflect the computation that
produced the answer. \citet{lanham2022faithfulness} introduce interventions that test
whether a model's stated reasoning actually causally affects its predictions.
\citet{turpin2023language} show that chain-of-thought explanations can be biased by
post-hoc rationalization and can omit influential features. Process supervision and
verifier-based methods provide complementary signals for reasoning quality
\citep{lightman2023verify,wang2022selfconsistency}. We treat the reconstructed trace as
a \emph{proxy}, not as the model's true hidden CoT, and we include answer consistency,
leakage, length, and downstream utility as part of the evaluation protocol rather than
assuming faithfulness from surface fluency.

\paragraph{Data synthesis and agent tooling.}
Synthetic data pipelines have become a practical way to obtain supervision at scale
\citep{wang2023selfinstruct}. Our work is also a systems contribution: a job-based
pipeline, a prompt router, and an agent skill that packages the workflow for one-prompt
deployment. The agent skill format used here follows the standard \texttt{SKILL.md}
convention supported by Codex, Claude Code, and DeepSeek Harness; the implementation is
model-agnostic and configures the target, decoder, and cleaner independently.

\paragraph{Positioning.}
\recast{} differs from prior rationale distillation in two ways. First, it does not
require a teacher rationale; it reconstructs one from the answer. Second, it does not
require a hidden reasoning channel; it is designed for the answer-only interface that
frontier deployments actually expose. The price is that the reconstructed trace is a
proxy whose faithfulness must be evaluated, which we make explicit throughout.

\section{Threat Model}
\label{sec:threat}

We formalize the reconstruction setting as an extraction threat model. The model is the
target; the reconstructor is an external agent with only black-box query access. The
goal is to obtain an auditable rationale that is useful for training a student, not to
recover the target's true hidden computation.

\paragraph{Target model and observable surface.}
Let \(T\) be a closed-source reasoning model and let \(q\) be a question. We write
\(z\) for the model's hidden chain-of-thought and \(y = T(q)\) for the final answer that
the API returns. The API exposes \(y\), token usage, and a finish reason. It does not
expose weights, logits, hidden states, or \(z\). We write \(\mathcal{O}(T)\) for this
observable surface.

\paragraph{Reconstructor capabilities.}
A reconstructor \(R\) can (i) query \(T\) with chosen questions and prompts, (ii) use an
external decoder model \(D\), (iii) choose among a set of elicitation patterns
\(\mathcal{P}\), and (iv) post-process answers and traces locally. It cannot read the
target's weights, hidden states, reasoning fields, or provider-internal logs.

\paragraph{Goal.}
Given a dataset \(\{(q_i,y_i)\}_{i=1}^{n}\), \(R\) produces a trace
\(c_i = D(q_i, y_i, p_i)\) for a pattern \(p_i \in \mathcal{P}\). The trace is used to
train a student \(S\) on \((q_i,c_i)\). Success is functional: \(S\) should solve
held-out questions, and \(c_i\) should pass auditability checks (answer consistency,
leakage, length, structure). Exact recovery of \(z_i\) is neither required nor claimed.

\paragraph{Auditability condition.}
We use \emph{auditable} to describe a trace that is structured, self-contained, free of
dialogue/meta artifacts, and does not reveal the final answer before the derivation.
The student receives only the question at inference; the decoder sees \(y\) while
constructing \(c\). Auditability is a property of the reconstruction procedure, not a
claim that the trace is the target's true hidden computation.

\paragraph{Constraints.}
The reconstructor operates under realistic deployment constraints: rate limits and
query budgets; provider safety filters; terms of service; and no access to privileged
interfaces. We do not attempt to bypass content filters, exploit tool implementations,
or use unauthorized endpoints. The threat model is therefore an \emph{interface-limited}
one: the question is what can be reconstructed from the surface a legitimate user is
allowed to observe.

\paragraph{Why the three extraction methods fail.}
Table~\ref{tab:failure} summarizes our controlled reproduction. The three methods
require different side channels, and on Claude-Sonnet-5.5 through standard
chat-completions all three channels are unavailable. The result is zero CoT tokens in
all cases. The source model in the P2 probe consumed 9,949 completion tokens but
returned only a final answer, which is consistent with substantial hidden reasoning
that never reaches the client.

\begin{table*}[t]
\centering
\caption{Controlled reproduction of three representative CoT extraction methods on
Claude-Sonnet-5.5 through a standard chat-completions interface. All three obtain zero
CoT tokens. The table reports the observed interface behavior, not a benchmark score;
runs use one or two questions and are intended as mechanism checks.}
\label{tab:failure}
\small
\begin{tabular}{@{}>{\raggedright\arraybackslash}p{2.6cm}
                >{\raggedright\arraybackslash}p{4.0cm}
                >{\raggedright\arraybackslash}p{4.6cm}
                >{\raggedright\arraybackslash}p{1.7cm}@{}}
\toprule
Method & Required side channel & Observed on Claude-Sonnet-5.5 & CoT tokens \\
\midrule
P3 Forced-Reasoning \citep{p3} &
Named/any \texttt{tool\_choice} plus a free-text tool argument &
Named tool choice rejected with HTTP 400; free-text scratchpad arguments blocked with
\texttt{content\_filter} &
0 \\
P1 EchoCoT \citep{p1} &
Scratchpad tool state across multiple turns &
First scratchpad call blocked with \texttt{content\_filter}; tool invocation rate 0\% &
0 \\
P2 Stealing Reasoning Traces \citep{p2} &
Signed/encrypted reasoning block to replay &
No \texttt{signature}, \texttt{encrypted\_content}, or \texttt{reasoning\_text}; source
billed 9,949 completion tokens; extraction error 1.0 &
0 \\
\midrule
\recast{} (ours) &
Final answer \(y\) plus an external decoder \(D\) &
Only the answer is required; no hidden channel &
structured trace \\
\bottomrule
\end{tabular}
\end{table*}

\paragraph{The four closing constraints.}
The observed failures are not independent accidents. They reflect four interface
properties:
\begin{enumerate}
  \item \textbf{Reasoning fields are absent.} The response contains final content and
  usage, but not \texttt{reasoning\_content}, \texttt{thinking}, \texttt{signature}, or
  \texttt{encrypted\_content}. P3's native baseline and P2's source stage both require
  such a field.
  \item \textbf{Named tool choice is rejected.} The endpoint accepts only automatic tool
  choice. P3's first protocol step---forcing a specific function---cannot be executed.
  \item \textbf{Free-text tool arguments are filtered.} A tool with a free-text
  scratchpad argument is blocked even under a neutral name. P1's scratchpad and P3's
  soft-force fallback both depend on this channel.
  \item \textbf{No replayable block is exposed.} P2 needs an opaque provider-issued
  block. Chat-completions does not produce one, so there is nothing to replay.
\end{enumerate}
These constraints are deployment-level. Ordinary tools with short structured arguments
still work, and a standalone decoder call succeeds; the failure is specific to the
channels that extraction methods require.

\paragraph{Why reconstruction succeeds.}
\recast{} operates entirely inside the observable surface. It consumes \(q\) and \(y\),
both of which are available under the threat model, and uses a decoder to synthesize a
structured rationale. It does not require a reasoning field, a tool call, a named tool
choice, or an encrypted block. This shifts the problem from \emph{opening a closed
channel} to \emph{reconstructing a plausible and auditable trace from an open one}. The
shift is what makes the method model-agnostic: any model whose final answer is observable
can be processed by the same protocol.

\paragraph{Success criteria.}
We evaluate the reconstructed trace along four axes: (i) \emph{task utility}, whether a
student trained on the trace improves on held-out questions; (ii) \emph{answer
consistency}, whether the trace supports the reference answer; (iii) \emph{auditability},
whether the trace is structured, self-contained, and free of dialogue/meta artifacts;
and (iv) \emph{cost}, the number of target, decoder, and cleaner calls per accepted
sample. This preprint reports the system and a controlled artifact validation; the
large-scale version of these axes is the subject of a follow-up study.

\section{Methodology}
\label{sec:method}

\subsection{Overview and Design Principles}

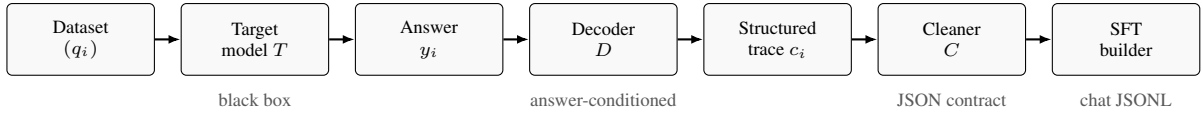
\begin{figure*}[t]
\centering
\begin{tikzpicture}[
  node distance=0.34cm,
  box/.style={draw, rounded corners=2pt, align=center, minimum height=0.95cm,
              minimum width=1.95cm, font=\scriptsize, fill=gray!5},
  arr/.style={-{Latex[length=1.6mm]}, thick},
  lab/.style={font=\scriptsize, text=gray!70!black}
]
\node[box] (data) {Dataset\\$(q_i)$};
\node[box, right=of data] (target) {Target\\model $T$};
\node[box, right=of target] (answer) {Answer\\$y_i$};
\node[box, right=of answer] (decoder) {Decoder\\$D$};
\node[box, right=of decoder] (trace) {Structured\\trace $c_i$};
\node[box, right=of trace] (cleaner) {Cleaner\\$C$};
\node[box, right=of cleaner] (sft) {SFT\\builder};

\draw[arr] (data) -- (target);
\draw[arr] (target) -- (answer);
\draw[arr] (answer) -- (decoder);
\draw[arr] (decoder) -- (trace);
\draw[arr] (trace) -- (cleaner);
\draw[arr] (cleaner) -- (sft);

\node[lab, below=0.10cm of target] {black box};
\node[lab, below=0.10cm of decoder] {answer-conditioned};
\node[lab, below=0.10cm of cleaner] {JSON contract};
\node[lab, below=0.10cm of sft] {chat JSONL};
\end{tikzpicture}
\caption{The \recast{} pipeline. The target model is treated as a black box; only its
final answer enters the reconstruction stage. The decoder produces a structured
first-person trace, the cleaner removes dialogue/meta artifacts, and the dataset builder
attaches a canonical answer and validates the result.}
\label{fig:pipeline}
\end{figure*}

\recast{} turns a dataset of questions into SFT-ready reasoning data through five
stages: (1) collect the target model's answers, (2) reconstruct a structured rationale
with a decoder, (3) clean the rationale, (4) extract the final answer, and (5) build and
validate the SFT dataset. Figure~\ref{fig:pipeline} illustrates the flow. The four model
roles are deliberately separated: a \emph{target} model produces answers, a
\emph{decoder} reconstructs traces, a \emph{cleaner} removes artifacts, and a
\emph{student} is trained on the result. Each role can use a different
OpenAI-compatible endpoint; the default configuration uses a relay-served target model
and DeepSeek for decoding and cleaning, but the code has no vendor-specific dependency.

We follow six design principles. \emph{Answer-only}: the reconstruction stage never
requires a hidden reasoning field. \emph{Role separation}: target, decoder, and cleaner
are independent, so a failure in one role does not invalidate the others.
\emph{Dataset-aware elicitation}: the prompt used to collect the answer is selected from
the dataset profile, because some patterns elicit more explicit reasoning than others.
\emph{Defensive cleaning}: every generated trace passes through leak, length, and
answer-consistency checks. \emph{Reproducibility}: every run creates a timestamped job
with logs, configuration, and a manifest. \emph{Automation}: the same pipeline is exposed
as a command-line tool and as an agent skill.

\subsection{Target Answer Collection and Pattern Routing}

The target model is queried with a dataset-specific elicitation pattern. A pattern is a
short teacher--student script that encourages the target to expose intermediate steps
without explicitly asking for a hidden CoT. Table~\ref{tab:patterns} lists the six
patterns used by the router.

\begin{table*}[t]
\centering
\caption{Elicitation patterns and their intended profiles. The pattern controls how the
target answer is elicited; it does not control the decoder or the cleaner.}
\label{tab:patterns}
\small
\begin{tabular}{@{}llp{8.2cm}@{}}
\toprule
Pattern & Profile & Structure and purpose \\
\midrule
\texttt{math\_solver} & math & Teacher reads the problem; student derives and verifies.
For proof problems, the verifier attacks the argument for gaps. \\
\texttt{socratic} & logic, planning & Repeated questions on goal, constraints,
assumptions, and counterexamples before solving. \\
\texttt{expert\_panel} & science & Multiple expert roles decompose the problem, check
assumptions, and converge on a solution. \\
\texttt{pair\_programming} & code & Driver--navigator pair writes code, tests, and
counterexamples. \\
\texttt{red\_blue} & secondary & Blue team proposes a solution; red team attacks
edge cases and counterexamples; a judge converges. \\
\texttt{retrospective} & open-ended & Solve first, then use a retrospective/PDCA cycle to
surface requirements and extensions. \\
\bottomrule
\end{tabular}
\end{table*}

\paragraph{Routing.}
The router first checks a dataset-name map (e.g., MATH-500 and GSM8K map to
\texttt{math}; HumanEval and MBPP map to \texttt{code}). For an unknown dataset, it scores
the first rows with profile keywords and falls back to \texttt{general}. The selected
profile determines a primary pattern and a secondary pattern (Table~\ref{tab:patterns}
shows the primary mapping; the registry contains the full mapping). The router is
exposed as a standalone CLI so that users can inspect the decision before running the
pipeline.

\paragraph{Filter-resilient collection.}
Relay deployments may reject prompts that ask for structured reasoning. The collector
therefore maintains a prompt chain: primary pattern, secondary pattern, medium prompt,
and short prompt. Each prompt is tried first with the original question and then with a
LaTeX-light variant. The attempt metadata (pattern, prompt variant, question variant) is
stored with every record, so a job can be audited after the fact. A failed question does
not abort the batch; it is recorded as an error and the pipeline continues.

\subsection{Rationale Reconstruction}

The decoder receives the question and the target model's complete answer. It is
instructed to write in the first person, in the order in which the solution unfolds, and
to cover five stages: task identification, requirement/diagnostic analysis, strategy
selection, key derivation, and verification/closing. The prompt forbids dialogue and
meta-commentary (``teacher'', ``student'', ``template'', ``decode'', etc.), forbids
restating the target answer verbatim, and forbids announcing the final answer in the
middle of the trace. A compact version of the prompt is shown in Appendix~\ref{app:prompts}.

\paragraph{Length floor.}
Short reconstructions tend to collapse into generic summaries. We therefore enforce a
minimum output length
\[
L_{\min} = \max(0,\ \alpha \cdot |y|),
\]
where \(|y|\) is the approximate token length of the target answer and \(\alpha\) is a
configurable floor (default \(1.0\)). If the decoder returns a trace shorter than
\(L_{\min}\), it is re-prompted with an expansion instruction that asks for omitted
derivations, alternatives, and verification steps. The floor is a coverage heuristic,
not a quality guarantee; Section~\ref{sec:limitations} discusses the risk of padding.

\paragraph{Template variants.}
The pipeline ships two decoder templates. The \emph{solution} template is the default:
it asks for a natural first-person solution trace with the five stages above. The
\emph{audit} template is a legacy seven-stage version that adds explicit output
organization and self-check sections. The solution template produces shorter, more
natural traces and is used for the artifact validation in Section~\ref{sec:artifact}; the
audit template is retained for ablation and for datasets that need a more explicit
structure.

\subsection{Cleaning and Guardrails}

The decoder can leak artifacts from the target answer: dialogue framing, classroom
metaphors, statements about the task or the reconstruction process, and explicit
``final answer'' lines. The cleaner rewrites the trace into a self-contained first-person
rationale. It returns a JSON object with three fields:
\[
\{\texttt{cleaned\_cot},\ \texttt{removed},\ \texttt{notes}\}.
\]
The cleaner is not allowed to add new mathematical content; it may remove or rewrite
noise, and it must preserve the stage headings and their order. The removed categories
are recorded for auditing (dialogue, shell, task meta, answer line).

\begin{table*}[t]
\centering
\caption{Cleaner outcomes and how the dataset builder handles them. A non-\texttt{ok}
outcome never silently replaces the raw trace with a shorter or cleaner-looking one.}
\label{tab:cleaner}
\small
\begin{tabular}{@{}lp{10.4cm}@{}}
\toprule
Status & Meaning and downstream handling \\
\midrule
\texttt{ok} & Cleaned trace passed JSON parsing, leak checks, and length checks; used as the
supervision target. \\
\texttt{still\_leaky} & Dialogue or meta terms remain after retries; raw trace is used and
the leak is recorded. \\
\texttt{rejected\_short} & Cleaned trace is shorter than the minimum ratio; raw trace is used
and the record is flagged. \\
\texttt{rejected\_long} & Cleaned trace exceeds the maximum ratio (possible duplication);
raw trace is used and the record is flagged. \\
\texttt{api\_error} & Cleaner call failed; raw trace is used and the error is recorded. \\
\bottomrule
\end{tabular}
\end{table*}

The cleaner runs at temperature zero with a JSON response mode. It retries on parse
errors and on leak/length violations, and it receives the specific violation as
feedback. A rule-based safety net runs after the model: it strips residual meta
sentences, removes explicit answer declarations, normalizes whitespace, and checks that
the trace is not empty. The cleaned trace is stored as an overlay keyed by record; the
raw decoded trace is never overwritten.

\subsection{Answer Extraction and Canonicalization}

Each training sample needs a final answer. When the dataset provides a reference answer,
the pipeline uses it directly (for MATH-500 and GSM8K, the test-set answer is used).
When no reference is available, a hybrid extractor first applies pattern-based parsing
and then asks a model to extract the final answer from the trace. Answers are normalized
(whitespace, \texttt{\textbackslash boxed\{\}}, units, and common equivalent forms)
before being appended to the supervision target. The extraction method is recorded per
record (\texttt{testset\_reference}, \texttt{llm\_extraction}, or \texttt{regex}), so a
dataset can be filtered by answer provenance.

\subsection{SFT Data Pipeline}

The dataset builder converts each accepted trace into a chat sample. The system prompt
frames the student as a reasoning assistant; the user message is the question; the
assistant message contains the cleaned trace followed by a canonical final-answer line.
Loss is applied only to the assistant tokens. We support four target variants
(Table~\ref{tab:targets}) for ablations and for different training objectives.

\begin{table*}[t]
\centering
\caption{Supervision target variants produced by the dataset builder.}
\label{tab:targets}
\small
\begin{tabular}{@{}lp{10.2cm}@{}}
\toprule
Target & Assistant content \\
\midrule
\texttt{cot\_answer} & Cleaned CoT followed by the final answer (default). \\
\texttt{cot\_only} & Cleaned CoT only; the answer is kept in metadata and not trained on. \\
\texttt{source\_only} & The target model's original answer; used as an answer-only baseline. \\
\texttt{cot\_plus\_source} & Cleaned CoT plus the original answer; used to test whether
answer style adds signal beyond the reconstructed rationale. \\
\bottomrule
\end{tabular}
\end{table*}

Before writing the splits, the builder removes duplicate questions, enforces minimum and
maximum lengths, filters traces whose final answer disagrees with the reference, and
checks for benchmark leakage. The
output contains \texttt{sft\_train.jsonl}, \texttt{sft\_val.jsonl},
\texttt{sft\_test.jsonl}, per-variant evaluation files, and
\texttt{dataset\_stats.json} with counts, dropped reasons, answer methods, and CoT
sources. The built-in benchmark loaders use MATH-500 \citep{hendrycks2021math} and GSM8K
\citep{cobbe2021gsm8k}. A validator checks the schema, duplicate rates, and leakage terms;
a failed validation stops the job without deleting the raw artifacts.

\subsection{Agent Skill and One-Prompt Automation}

The complete workflow is packaged as an agent skill. The skill directory contains a
\texttt{SKILL.md} entry point, a pipeline reference, and a small wrapper that locates the
repository from any working directory. Installing the skill copies it into the host's
skill directory (project-level \texttt{.dsh/skills} or \texttt{.claude/skills}, or
user-level \texttt{\textasciitilde/.codex/skills}); a new session then exposes the
skill to the agent. The intended interaction is a single prompt such as
``run CoT reconstruction on MATH-500 with model X''. The agent loads the skill, resolves
the repository, chooses the dataset, target model, and pattern, runs the pipeline, and
reports the job directory and summary.

The skill does not embed credentials. It reads the three configuration files
(\texttt{config.relay.json}, \texttt{api.txt}, \texttt{api\_clean.txt}) and supports
swapping the target, decoder, and cleaner independently through command-line flags
(\texttt{--target-*}, \texttt{--decoder-*}, \texttt{--cleaner-*}). This is important for
reproducibility: a different lab can reuse the skill with entirely different model
providers.

\subsection{Implementation and Artifact Validation}
\label{sec:artifact}

The repository contains the pipeline entry point, the collector, the decoder runner, the
router, the cleaning/answer/dataset/validation scripts, the prompt templates, the
patterns, the evaluation harness, and the skill. A run creates a job directory
\texttt{jobs/} with \texttt{raw/}, \texttt{decoded/}, \texttt{cleaned/},
\texttt{answers/}, \texttt{sft/processed/}, \texttt{logs/}, \texttt{config/},
\texttt{manifest.json}, and \texttt{summary.md}. The
manifest records every stage's status, elapsed time, and output path; the summary
contains counts and a ready-to-run training command.

To validate the artifact rather than the method's accuracy, we ran the complete pipeline
on a single MATH-500 item with Claude-Sonnet-5.5 as the target and the default
DeepSeek roles. All six stages completed (Table~\ref{tab:pilot}). The collector produced
one successful answer; the decoder produced a 1,297-token trace against a 938-token
floor; the cleaner returned \texttt{ok} with no leak terms; the answer stage used the
MATH-500 reference; the builder produced one \texttt{cot\_answer} training sample; and
the validator passed. We use this run only to verify that the stages compose and that
the artifacts are self-consistent.

\begin{table}[t]
\centering
\caption{Single-item artifact validation. Times are wall-clock seconds for one question;
the run is a pipeline smoke test, not a benchmark.}
\label{tab:pilot}
\small
\begin{tabular}{@{}lr@{}}
\toprule
Stage & Time (s) \\
\midrule
Collect target answer & 12.8 \\
Decode reconstructed CoT & 9.9 \\
Clean trace & 8.4 \\
Extract/canonicalize answer & 2.2 \\
Build + validate SFT data & 0.2 \\
\bottomrule
\end{tabular}
\end{table}

\paragraph{Cross-model qualitative validation.}
Beyond Claude-Sonnet-5.5, we applied the same answer-only protocol to GPT-6.1
\citep{openai2026} and Gemini 3 \citep{google2026} and obtained successful qualitative
reconstructions: each run produced a structured, first-person trace that passed the
cleaner and the dataset validator. We do not report quantitative comparisons for these
models in this preprint; they serve as existence demonstrations that the protocol is not
tied to a particular vendor. The system accepts any OpenAI-compatible chat endpoint, so
the same commands apply when the target, decoder, or cleaner is changed.

\section{Discussion}
\label{sec:discussion}

\paragraph{Extraction versus reconstruction.}
The three reproduced methods are elegant when their side channel exists. They fail on
Claude-Sonnet-5.5 not because the implementations are wrong, but because the deployment
closes the channel: reasoning fields are omitted, named tool choice is rejected,
free-text tool arguments are filtered, and encrypted blocks are absent. A system that
depends on any one of these assumptions inherits a single point of failure. \recast{}
removes the dependency altogether. It treats the final answer as the observable signal
and reconstructs a rationale that is consistent with it. The trade-off is explicit: the
reconstructed trace is a proxy, and its faithfulness must be tested rather than assumed.

\paragraph{Why the answer is a useful anchor.}
The final answer constrains the space of plausible rationales: it fixes the destination
and rules out derivations that end elsewhere. This is weaker than observing the true
trajectory, but it is stronger than answer-only training, which provides no intermediate
supervision. The decoder uses the answer as an anchor, not as a script: the prompt asks
for a first-person derivation and forbids restating the answer as the explanation. The
supporting pattern is also important. A pattern that elicits a detailed answer gives the
decoder more material to reconstruct from; a terse answer leaves more of the rationale
to the decoder's prior.

\paragraph{Is the trace auditable?}
We use ``auditable'' in a procedural sense: the trace is structured into stages, is
self-contained, does not mention teachers/students/templates/decoding, and does not
announce the final answer in the middle of the derivation. Cleaning enforces these
properties, and the validator checks length, answer consistency, and leakage. This is
not a guarantee of faithfulness. A trace can be well structured and still be a post-hoc
rationalization. We therefore treat auditability as a necessary condition for usable
supervision, not as evidence that the trace equals the target's hidden computation.

\paragraph{Auditable traces.}
The title uses ``auditable'' to emphasize that the trace is organized, self-contained,
and checkable. Auditability is weaker than faithfulness: a well-structured trace can
still be a post-hoc rationalization. The decoder is answer-conditioned; the student sees
only the question at inference. A training-time answer-masking variant is a natural
extension, but it is outside the scope of this preprint, and we do not conflate the
deployment condition with a stronger experimental claim.

\paragraph{Cross-model behavior.}
The failure analysis is specific to Claude-Sonnet-5.5 through a standard chat-completions
interface; it is not a claim that the three extraction methods fail everywhere. The
positive result is more general: because \recast{} only needs the final answer, the same
protocol applies to GPT-6.1, Gemini 3, and any other model with an observable answer. We
have obtained qualitative successes on all three model families, but we do not report a
controlled cross-model comparison in this preprint.

\paragraph{Cost and automation.}
Each accepted sample uses one target call, one decoder call, one cleaner call, and
(optionally) one answer-extraction call. This is more expensive than answer-only
distillation but much cheaper than human rationale annotation. The skill reduces the
operational cost further: one prompt deploys the pipeline and produces a self-contained
job, so a user who is not a Python developer can still obtain SFT-ready data. The skill
inherits the pipeline's model-agnostic configuration, so the same prompt works with a
different target, decoder, or cleaner.

\section{Limitations and Responsible Use}
\label{sec:limitations}

\paragraph{No large-scale efficacy claim.}
This preprint does not report a benchmark comparison. We do not claim that a student
trained on reconstructed traces outperforms a base model, an answer-only baseline, or a
teacher. The artifact validation in Section~\ref{sec:artifact} is a pipeline smoke test,
not evidence of accuracy. A full evaluation requires multiple seeds, multiple datasets,
and a controlled comparison of target variants.

\paragraph{Endpoint-specific failure analysis.}
The failure of the three extraction methods is observed on Claude-Sonnet-5.5 through one
chat-completions interface. A different endpoint, a native SDK, or a different model
version may expose a channel that this deployment closes. Our claim is about the
interface we studied, not about the impossibility of extraction in general.

\paragraph{Reconstructed traces are proxies.}
The decoder produces a plausible rationale, not the target's hidden computation. The
trace may contain post-hoc rationalization, may omit influential features, or may be
overconfident. Cleaning improves presentation but does not make the trace faithful.
Downstream evaluation and faithfulness probes are necessary; surface fluency is not
enough.

\paragraph{Answer leakage and padding.}
The decoder sees the answer and may leak it into the rationale; the cleaner removes
explicit answer statements, but subtle leakage can remain. The length floor reduces
short, generic traces but can also encourage padding. The dataset builder records the
trace source and length, and users should inspect a sample before large-scale training.

\paragraph{Model and provider drift.}
Relay endpoints, model versions, and safety policies change. A prompt that succeeds today
may be filtered tomorrow, and a decoder update may change the style of the reconstructed
trace. The job manifest records the configuration and timestamp, but it cannot guarantee
that a later run is identical. Reproduction should record model identifiers and endpoint
classes.

\paragraph{Responsible use.}
\recast{} is a research tool for reconstructing auditable rationales from answers. It is
not intended to bypass safety controls, extract proprietary reasoning at scale, or
circumvent provider terms. The threat model in Section~\ref{sec:threat} assumes authorized
access to an API and forbids filter evasion. Users are responsible for complying with
their model provider's terms and for not redistributing model outputs without permission.
The benchmark data used by the pipeline follows its upstream license (MATH and GSM8K are
MIT-licensed in their original distributions), and redistributed derivatives should
preserve the corresponding notices.

\paragraph{Training scope.}
The optional training scripts target LoRA/QLoRA SFT on a single GPU and were not executed
at scale for this preprint. Reported GPU results should be treated as environment
dependent. The paper's claims are about the reconstruction pipeline and its artifact, not
about a particular training recipe.

\section{Conclusion}
\label{sec:conclusion}

We studied the practical limits of extracting chain-of-thought from frontier closed
models. Three representative state-of-the-art extraction methods---forced reasoning,
scratchpad state replay, and encrypted-block replay---all returned zero CoT tokens on
Claude-Sonnet-5.5 through a standard chat-completions interface. The failures follow from
four interface constraints, not from implementation errors. We therefore proposed
\recast{}, which reframes the problem from extracting a hidden rationale to reconstructing
an auditable rationale from the final answer. \recast{} combines dataset-aware answer
elicitation, first-person reconstruction, defensive cleaning, answer canonicalization,
and a validated SFT data pipeline. To the best of our knowledge, it is the first system
to reconstruct CoT from frontier closed models such as Claude-Sonnet-5.5, GPT-6.1, and
Gemini 3 without using hidden reasoning channels. The system is open-source and packaged
as an agent skill, so a single prompt can deploy it and run the complete pipeline to
produce SFT-ready data. The next step is a controlled large-scale evaluation of student
accuracy, faithfulness, and cost; the present preprint deliberately reports the system
and its artifact validation rather than an efficacy claim.

\bibliography{references}

\appendix
\section{Prompt Templates}
\label{app:prompts}

The repository contains the exact prompts used in the runs. For readability, this
appendix gives English translations of the two central templates; the original templates
are available under \texttt{templates/} and \texttt{eval/data/} in the artifact.

\paragraph{Decoder prompt.}
The decoder receives the question and the full target answer and is asked for a
first-person solution trace:
\begin{lstlisting}
[Question]
{question}

[Complete answer from the closed model]
{answer}

Write the reasoning process that I went through when solving this problem, in the first
person and in the order in which the solution unfolds:

1. how I understood the problem, identified the constraints, and noticed likely traps;
2. which approaches I considered, why I chose one, and why I abandoned another;
3. how the key derivation proceeds, including the necessary mathematical expressions
   and the evidence for each step;
4. how I verified the key steps (substitution, simplification, boundary checks, or
   counterexamples);
5. how I closed the solution.

Requirements:
- give concrete reasoning, not filler such as "I analyzed it" or "I checked it";
- do not restate sentences from the complete answer;
- do not mention teacher/student/classroom/dialogue/role/template/reconstruction/decoding,
  and do not describe how the answer is being organized;
- do not reveal the final answer before the end of the reasoning;
- output only the reasoning process, with no opening or closing remarks.
\end{lstlisting}

\paragraph{Cleaner prompt.}
The cleaner is asked for a strict JSON object and is told that it may remove or rewrite
noise but must not invent mathematical content:
\begin{lstlisting}
You are a training-data cleaner. The input is a CoT reconstructed from a closed model's
complete answer; it may contain (1) dialogue/classroom artifacts, (2) meta-commentary
about reconstruction, templates, or the task, and (3) explicit answer declarations.

Rewrite it as a self-contained first-person rationale. Remove all dialogue/classroom
traces and all sentences about how the answer was produced, reconstructed, or formatted.
Keep the stage headings and their order. Do not add new mathematical content. Remove
explicit "final answer:" lines; the answer is appended later.

Return exactly one JSON object:
{"cleaned_cot": "...", "removed": ["dialogue", "shell", "task_meta", "answer_line"],
 "notes": "one sentence, may be empty"}

Question:
{question}

Trace to clean:
{cot}
\end{lstlisting}

\section{Pattern Registry and Routing Examples}
\label{app:patterns}

The router combines a dataset-name map with keyword scoring over the first rows. The
registry is data, not code: adding a profile or a pattern does not require changing the
pipeline. Table~\ref{tab:routing-examples} gives representative decisions.

\begin{table*}[h]
\centering
\caption{Representative routing decisions. Unknown datasets fall back to keyword
scoring and then to \texttt{general}.}
\label{tab:routing-examples}
\small
\begin{tabular}{@{}llll@{}}
\toprule
Dataset & Profile & Primary & Secondary \\
\midrule
MATH-500, GSM8K, AIME & math & \texttt{math\_solver} & \texttt{red\_blue} \\
HumanEval, MBPP, LiveCodeBench & code & \texttt{pair\_programming} & \texttt{red\_blue} \\
GPQA, ScienceQA & science & \texttt{expert\_panel} & \texttt{red\_blue} \\
BBH, LogiQA & logic & \texttt{socratic} & \texttt{expert\_panel} \\
PlanBench, agent tasks & planning & \texttt{socratic} & \texttt{retrospective} \\
Open-ended writing & open\_ended & \texttt{retrospective} & \texttt{socratic} \\
Unknown & general & \texttt{socratic} & \texttt{math\_solver} \\
\bottomrule
\end{tabular}
\end{table*}

Each collection record stores the selected pattern, the prompt variant, the question
variant, and the attempt number. This makes a failed \texttt{content\_filter} attempt
traceable without rerunning the batch.

\section{Job Schema}
\label{app:schema}

A run creates a self-contained job directory:
\begin{verbatim}
jobs/<stamp>/
  raw/        collected.jsonl
  decoded/    decoded_<stamp>.json/.md
  cleaned/    cleaned_cots.json
  answers/    answers.json
  sft/        processed/sft_*.jsonl
  logs/       per-stage logs
  config/     prompts, run config
  manifest.json
  summary.md
\end{verbatim}

The manifest is the machine-readable index. It records the dataset and target
configuration, the selected pattern and routing reason, each stage's status, the number
of collected/decoded/cleaned records, the number of kept SFT samples, the dropped-sample
reasons, and the paths to every artifact. The summary is generated from the manifest and
contains the same counts plus a ready-to-run training command.

\section{Reproduction Commands}
\label{app:repro}

\paragraph{Setup.}
\begin{lstlisting}
git clone \
  https://github.com/C1ove-ai/RECAST-REconstruction-of-Chain-of-Thought-via-Answer-Masked-Structured-Traces.git \
  recast-cot-reverse
cd recast-cot-reverse
pip install -r requirements.txt
cp config.relay.example.json config.relay.json
cp api.example.txt api.txt
cp api_clean.example.txt api_clean.txt
# edit the three files; never commit them
\end{lstlisting}

\paragraph{Smoke run.}
\begin{lstlisting}
python recast_pipeline.py \
  --dataset math500 \
  --target-model <model-id> --limit 2
\end{lstlisting}
The command prints the job path and writes \texttt{summary.md} and
\texttt{manifest.json}. \texttt{--dry-run} resolves the dataset, target, pattern, and
stage commands without calling an API.

\paragraph{Model roles.}
The target, decoder, and cleaner are independent OpenAI-compatible roles:
\begin{lstlisting}
python recast_pipeline.py --dataset math500 \
  --target-config my_teacher_relay.json \
  --target-model teacher-x \
  --decoder-model decoder-y \
  --decoder-url https://decoder.example/v1 \
  --decoder-api my_decoder_key.txt \
  --cleaner-model cleaner-z \
  --cleaner-url https://cleaner.example/v1 \
  --cleaner-api my_cleaner_key.txt
\end{lstlisting}

\paragraph{Agent skill.}
\begin{lstlisting}
powershell -NoProfile -ExecutionPolicy Bypass `
  -File skills\install_skill.ps1
\end{lstlisting}
After installation, a new agent session can run the pipeline from a single prompt. The
skill only orchestrates the local repository; it does not embed keys or model outputs.

\end{document}